\documentclass{article} 
\usepackage{iclr2022_conference_arxiv,times}
\usepackage{defs_ahn}

\usepackage{hyperref}
\usepackage{graphicx}
\usepackage{caption}
\usepackage{subcaption}
\usepackage{mathtools}
\usepackage{amsmath, amsthm, amssymb}
\usepackage{url}
\usepackage{lipsum}
\usepackage{xargs}
\usepackage{xcolor}
\usepackage{enumitem}
\usepackage[title]{appendix}
\usepackage[export]{adjustbox}
\usepackage{algorithm2e}
\usepackage[font=small,labelfont=bf]{caption}

\usepackage{wrapfig}
\usepackage{multirow}
\usepackage{booktabs}

\usepackage[colorinlistoftodos,textsize=tiny]{todonotes}
\newcommandx{\sj}[2][1=]{\todo[linecolor=blue,backgroundcolor=blue!5,bordercolor=blue,#1]{#2}}
\newcommandx{\gautam}[2][1=]{\todo[linecolor=orange,backgroundcolor=orange!5,bordercolor=orange,#1]{#2}}

\makeatletter
\def\thanks#1{\protected@xdef\@thanks{\@thanks
        \protect\footnotetext{#1}}}
\makeatother

\title{TransDreamer: Reinforcement Learning with\\ Transformer World Models}

\author{Chang Chen$^{1\dag}$, Yi-Fu Wu$^{1\dag}$, Jaesik Yoon$^{1\dag}$, Sungjin Ahn$^{1, 2}\thanks{$^\dag$Equal contribution. \texttt{\{chang.chen,jaesik.yoon,yifu.wu,sjn.ahn\}@rutgers.edu}}$ \\
$^1$Rutgers University \& $^2$KAIST
}

\iclrfinalcopy 
\begin{document}

\maketitle

\begin{abstract}
The Dreamer agent provides various benefits of Model-Based Reinforcement Learning (MBRL) such as sample efficiency, reusable knowledge, and safe planning. However, its world model and policy networks inherit the limitations of recurrent neural networks and thus an important question is how an MBRL framework can benefit from the recent advances of transformers and what the challenges are in doing so. In this paper, we propose a transformer-based MBRL agent, called TransDreamer. We first introduce the Transformer State-Space Model, a world model that leverages a transformer for dynamics predictions. We then share this world model with a transformer-based policy network and obtain stability in training a transformer-based RL agent. In experiments, we apply the proposed model to 2D visual RL and 3D first-person visual RL tasks both requiring long-range memory access for memory-based reasoning. We show that the proposed model outperforms Dreamer in these complex tasks.

\end{abstract} 

\section{Introduction}
Model-based reinforcement learning (MBRL)~\citep{dyna} provides a solution for many problems of current reinforcement learning. Its imagination-based training provides sample efficiency by fully leveraging the interaction experience via the world model~\citep{worldmodels}. The world model can be considered a form of task-agnostic general knowledge enabling reusability of the knowledge about the environment in many downstream tasks~\citep{plan2explore}, and finally, the dynamics model makes planning possible~\citep{muzero,planet} for accurate and safe decisions~\citep{safe_mbrl,morel,lisp}. 

Among the recent advances in MBRL, a particularly notable one is the Dreamer agent~\citep{dreamer,dreamerv2}. Learning a world model in latent representation space, Dreamer is the first visual MBRL model that achieves performance and sample efficiency better than model-free approaches such as Rainbow~\citep{rainbow} and IQN~\citep{iqn}. 

To deal with partial observability~\citep{pomdp}, the dynamics models in MBRL have been implemented using recurrent neural networks (RNNs)~\citep{dreamer,dreamerv2,muzero,simple}. However, Transformers~\citep{transformer,trxl} have shown to be more effective than RNNs in many domains requiring long-term dependency and direct access to memory for a form of memory-based reasoning~\citep{ritter2020rapid,memo}. Also, it has been shown that training complex policy networks based on transformers using only rewards is difficult~\citep{gtrxl}, so learning a transformer-based world model where the training signal is more diverse may facilitate learning. Therefore, it is important to investigate how to make an MBRL agent using a transformer-based world model and to analyze the benefits and challenges in doing so.

In this paper, we propose a transformer-based MBRL agent, called TransDreamer. As implied by the name, the proposed model inherits from the Dreamer framework, but aims to bring the benefits of transformers into it.
Seemingly, it may seem like a simple plug-in task to replace an RNN with a transformer.
However, there are a few critical challenges to make it work.
First of all, we need to develop a new transformer-based world model that supports effective stochastic action-conditioned transitions in the latent space to implement the prior and posterior of the transition model. At the same time, this model should also preserve the parallel trainability of transformers for computational efficiency. To the best of our knowledge, there is no such model yet. 
Also, as shown in \cite{gtrxl}, finding an architecture, hyperparameters, and other design choices to make a working model is particularly challenging for Transformer-based RL models. 
 
The main contribution of this paper is the first transformer-based MBRL agent. We introduce the Transformer State-Space Model (TSSM) as the first transformer-based stochastic world model. Using this world model in the Dreamer framework, we propose TransDreamer, a fully transformer-based MBRL framework.
In experiments, we show that TransDreamer outperforms Dreamer on tasks that require long-term and complex memory interactions, and the world model of TransDreamer is better than Dreamer at predicting rewards and future frames for imagination.
Furthermore, we also show that the performance of TransDreamer is comparable to Dreamer on a few simple DMC \citep{dmcontrol} and Atari \citep{atari_env} tasks that do not require long-term memory.

\section{Preliminaries}
Our model builds on top of Dreamer~\citep{dreamer,dreamerv2}, a model-based reinforcement learning framework~\citep{dyna} for visual control in a partially observable Markov decision process (POMDP)~\citep{pomdp}.
Dreamer consists of three main steps: (1) world model learning, (2) policy learning, and (3) environment interaction. These steps are cycled until convergence. Specifically, a dynamics model of the environment (i.e., world model) and a reward function are learned to fit the data in an experience replay buffer. Then, an actor-critic policy is trained on imagined experience, i.e., hypothetical trajectories generated by simulating the learned world model. Lastly, to provide fresh experiences to the replay buffer, the agent collects trajectory data by executing the trained policy in the real environment.

\subsection{World Model in Dreamer}
The backbone of the world model used in Dreamer is a stochastic recurrent neural network, called the Recurrent State-Space Model (RSSM) \citep{planet}. Depicted in  Figure \ref{fig:rssm}, the RSSM represents a latent state $s_t$ by the concatenation of a stochastic state $z_t$ and a deterministic state $h_t$ which are updated by $z_t \sim p(z_t|h_t)$ and $h_t=f(h_\tmo,z_\tmo,a_\tmo)$, respectively.
Here, $a_\tmo$ is an action and the deterministic update $f$ is the state update rule of a recurrent neural network (RNN) such as an LSTM~\citep{lstm} or GRU~\citep{gru}. While the deterministic path helps to model the temporal dependency in the world model, it also inherits the limitations of RNNs, particularly when compared to the benefits of transformers. The stochastic state makes it possible to capture the stochastic nature of the world, e.g., for imagining multiple hypothetical future trajectories. Crucially, using the above models, rollouts can be executed efficiently in a compact latent space without the need to generate observation images.

Learning the RSSM is via the maximization of evidence lower bound~\citep{intro_variational}. The representation model $z_t \sim q(z_t|h_t,x_t)$ infers the stochastic state given an observation $x_t$. Whenever a new observation is provided, the current state is updated by the representation model. The observation model $p(x_t|s_t)$ and the reward model $p(r_t|s_t)$ are then used for the reconstruction of observation $x_t$ and reward $r_t$ from the latent state. All component models are listed in Table~\ref{table:rssm_tssm}.

\subsection{Policy Learning in Dreamer}\label{ssec:dreamer_policy}
After updating the world model for a number of iterations, Dreamer updates its policy $\pi_\phi(a_t|s_t)$. The policy learning is done without interaction with the actual environment; it uses imagined trajectories obtained by simulating the learned world model. Specifically, from each state $s_t$ obtained from  a batch sampled from the replay buffer, it generates a future trajectory of length $H$ using the RSSM world model and the current policy as the behavior policy for the imagination. Then, for each state in the trajectory, the rewards $p_\ta(r_t|s_t)$ and the values $v_\psi(s_t)$ are estimated. This allows us to compute the value estimate $V(s_t)$, e.g., by the discounted sum of the predicted rewards and the bootstrapped value $v(s_{t+H})$ at the end of the trajectory. See \cite{dreamer} for more details and other options about the value estimation.

Learning the policy in Dreamer means updating two models, the policy $\pi_\phi(a_t|s_t)$ and the value model $v_\psi(s_t)$. For updating the policy, Dreamer uses the sum of the value estimates of the simulated trajectories, $\smash{\sum_{\tau=t}^{t+H} V(s_\tau)}$, to construct the objective function. While we can compute the gradient of this objective w.r.t.~the parameters of the policy $\phi$ via a policy gradient method such as REINFORCE~\citep{reinforce}, Dreamer also takes advantage of the differentiability of the learned world model by directly backpropagating from the value function to the world model, and to the parameters of the policy network. This provides gradients of lower variance than that of REINFORCE. Updating the value model parameter $\psi$ is done via temporal difference learning with the value estimate $V(s_t)$ as the target. 

\section{TransDreamer}

Transformers have been shown to outperform RNNs in many tasks in both NLP and computer vision. In particular, their ability to directly access historical states and to learn complex interactions among them, has been shown to excel in tasks that require complex long-term temporal dependencies such as memory-based reasoning~\citep{ritter2020rapid,memo}. Furthermore, they have been shown to be effective for temporal generation in both language and visual domains. Observing that both of these abilities are important and desirable properties for a robust world model, we hypothesize that a model-based agent based on transformers can outperform the RNN-based Dreamer agent for tasks requiring complex and long-term memory dependency. 




\begin{figure}[t]
    \vspace{-2mm}
    \centering
    \begin{subfigure}[b]{0.33\textwidth}
         \centering
         \includegraphics[width=\textwidth]{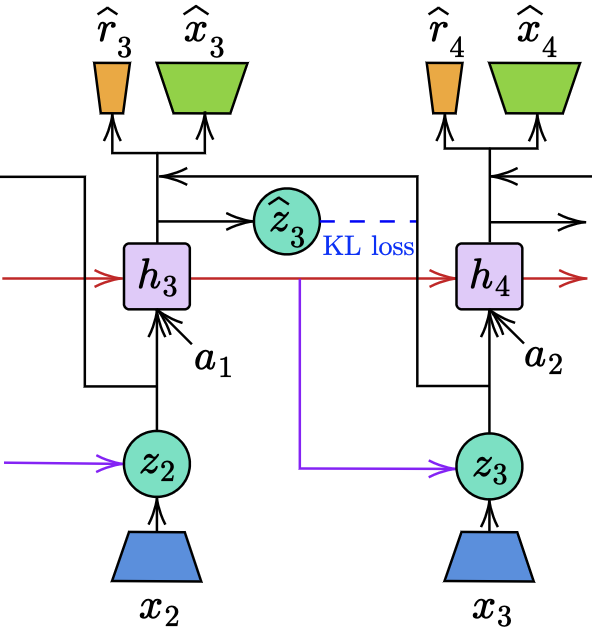}
         \caption{RSSM}
         \label{fig:rssm}
     \end{subfigure}
     \hspace{25mm}
     \begin{subfigure}[b]{0.345\textwidth}
         \centering
         \includegraphics[width=\textwidth]{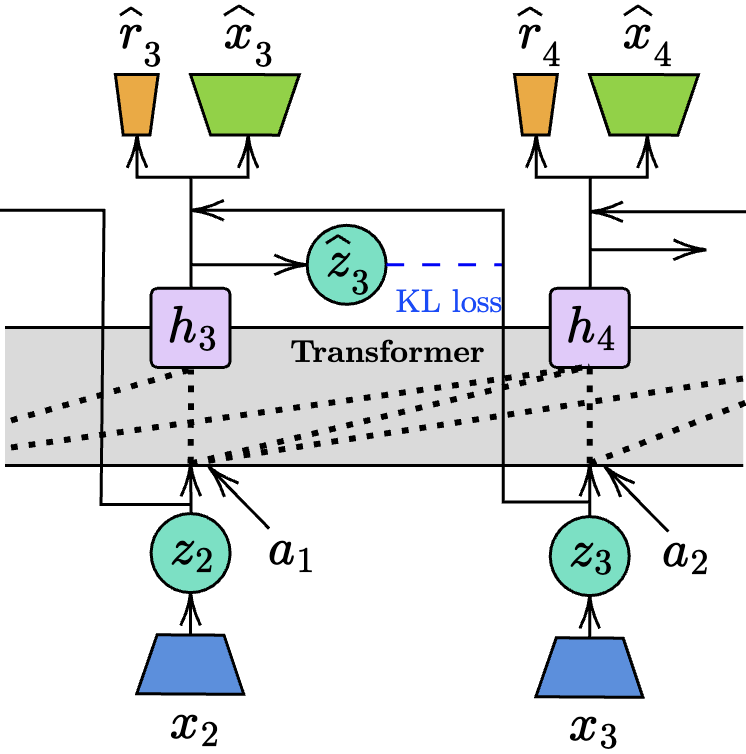}
         \caption{TSSM}
         \label{fig:tssm}
     \end{subfigure}
     \vspace{-2mm}
    \caption{RSSM and TSSM. The red arrow in RSSM makes sequential computational necessary. In TSSM, we replace this by a transformer. In addition, the purple arrow should also be removed in TSSM because it also prevents parallelizing the updates of all time steps.}
    \vspace{-6mm}
    \label{fig:rssm_tssm}
\end{figure}

\subsection{Transformer State Space Model (TSSM)}

In the design of a transformer-based world model, we aim to achieve the following desiderata: (i) to directly access past states, (ii) to update the states of each time step in parallel during training, (iii) to be able to roll out sequentially for trajectory imagination at test time, and (iv) to be a stochastic latent variable model. To our knowledge, no such model is available. The RSSM does not satisfy (i) and (ii). Although we can consider a simple modification of the RSSM, a memory-augmented RSSM, by introducing direct attention to the past states in the RNN state update~\citep{ke2018sparse} and make (i) satisfied as well, it still does not satisfy (ii) and thus remains computationally inefficient. Traditional transformers are deterministic and thus do not satisfy (iv). This motivates us to introduce the Transformer State-Space Model (TSSM).

The TSSM is a stochastic transformer-based state-space model. Figure~\ref{fig:rssm_tssm} illustrates the architectures of TSSM in comparison to RSSM. In the RSSM, the main source of the sequentially-dependent computation is the RNN-based state update $h_t = f_\text{gru}(h_\tmo,z_\tmo,a_\tmo)$, depicted in red arrows in Figure~\ref{fig:rssm}. This means that all component models of the RSSM are computed sequentially because they all take the hidden state as input. To remove this sequential computation and enable direct access to and complex interaction of the historical states, we propose employing a transformer as a replacement for the RNN.
Unlike the RSSM which accesses the past indirectly only via a compression $h_\tmo$, the transformer is allowed to directly access the sequence of stochastic states and actions of the past at every time step, i.e., $h_t = f_\text{transformer}(z_{1:\tmo}, a_{1:\tmo})$. If all $h_t$ can be computed in parallel in this way, then all components taking $h_t$ as input can also be computed in parallel.

\textbf{Myopic Representation Model}. This hope, however, is broken due to the representation model $q(z_t|h_t,x_t)$. This is because unlike other component models listed in Table~\ref{table:rssm_tssm}, the output of the representation model is used as the input to the transformer. Since a transformer should not use an output also as an input to achieve parallel training, the representation model should not be conditioned on $h_t$. That is, the purple arrows in Figure~\ref{fig:rssm_tssm} should be removed. To this end, we propose approximating the posterior representation model by $q(z_t|x_t)$, removing $h_t$.
Since the posterior can now be computed independently for each time step, we can compute all of the inputs $z_{1:\tmo}$ simultaneously.
Then, with a single forward pass through the transformer, we can obtain $h_{1:t}$.

\begin{table}[t]
\vspace{-3mm}
\caption{Comparison of the Component Models of RSSM and TSSM.}
\vspace{-2mm}
\label{table:rssm_tssm}
\centering
\begin{tabular}{l|c|c}
\multicolumn{1}{c|}{}  
& \textbf{RSSM} 
& \textbf{TSSM} \\ \hline\hline
{{Deterministic state model}} 
& $h_t = \text{gru}(h_\tmo,z_\tmo,a_\tmo)$ 
& $h_t = \text{transformer}(z_\ottmo, a_\ottmo)$ \\ \hline
{{Representation model}} 
& $z_t \sim q(z_t|h_t,x_t)$ 
& $z_t \sim q(z_t|x_t)$  \\ \hline
Stochastic state model 
& \multicolumn{2}{c}{$\hat{z}_t \sim p(\hat{z}_t|h_t)$} \\\hline
Image predictor 
& \multicolumn{2}{c}{$\hat{x}_t \sim p(\hat{x}_t|h_t,z_t)$} \\\hline
Reward predictor 
& \multicolumn{2}{c}{$\hat{r}_t \sim p(\hat{r}_t|h_t,z_t)$} \\\hline
Discount predictor 
& \multicolumn{2}{c}{$\hat{\ga}_t \sim p(\hat{\ga}_t|h_t,z_t)$} \\\hline
\end{tabular}
\vspace{-3mm}
\end{table}

\begin{wrapfigure}{r}{0.35\textwidth}
    \vspace{-7mm}
    \centering
    \includegraphics[width=0.35\textwidth]{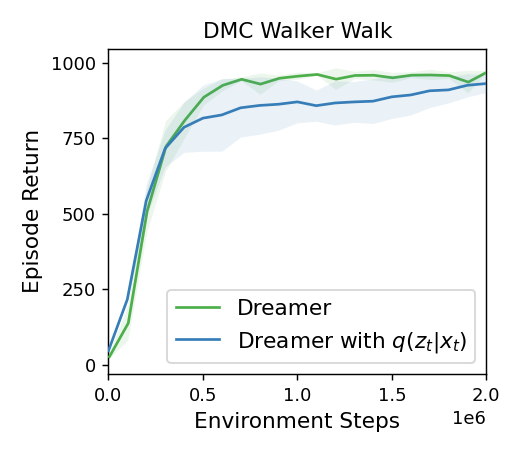}
    \vspace{-7mm}
    \label{fig:static_post}
\end{wrapfigure}

One may argue that removing $h_t$ and thereby ignoring all the history $z_{1:\tmo}$ in the representation model may result in a poor approximation. This can be true if we use only $z_t$ as the full state of our model. However, like the RSSM, the full state of our model is a concatenation of the stochastic state $z_t$ and the deterministic state $h_t$. Therefore, we hypothesize that encoding temporal information in the stochastic states may not be essential for model performance because that information is provided by the deterministic states. Furthermore, the trajectory imagination does not require the representation model but only requires the stochastic state model $p(z_t|h_t)$ that can still use $h_t$. We observe that if this hypothesis is correct, a modified Dreamer using $q(z_t|x_t)$ instead of $q(z_t|x_t,h_t)$ would perform similarly to the original Dreamer and the plot on the right confirms this. Another possible yet more complex choice for the representation model is to condition the posterior representation directly on all the past observations using another transformer layer, i.e.,  $q(z_t|f_\text{transformer}(x_{1:t}, a_{1:t}))$. However, due to its increased complexity, we do not consider this model.

\textbf{Imagination}. 
During imagination, we use the prior stochastic state $\hat{z}_t\sim p(\hat{z}|h_t)$ as the input to the transformer to autoregressively generate future states as shown in Figure~\ref{fig:rollout} in the Appendix.
This allows the agent to imagine future states completely in the latent space but with more direct access to the historical states than the RSSM. In this way, the TSSM achieves all the desiderata discussed above. Table \ref{table:rssm_tssm} highlights the key differences between the RSSM and the TSSM.

The \textbf{loss function} is almost the same as that of the RSSM. The difference is that we approximate the representation posterior $p(z_{1:t}|x_{1:t})$ by $\smash{\pd{\tau}{t}q_\phi(z_\tau|x_\tau)}$ instead of $\smash{\pd{\tau}{t}q_\phi(z_\tau|z_{1:\tau-1},x_{1:\tau})}$ used in the RSSM.
The loss function can be found in Appendix \ref{appx:sec_loss_func} with the derivation of the ELBO.




\subsection{Policy Learning and Implementation Details}


\textbf{Policy Learning.} The policy learning in TransDreamer inherits the general framework of Dreamer described in Section~\ref{ssec:dreamer_policy}. The main difference is that we replace the RSSM with the TSSM. The component models are thus based on the states from the TSSM which can capture the long-term and complex temporal dependency better. Since the TSSM is fully differentiable, we can similarly use both REINFORCE and dynamics backpropagation to train the policy.
The TSSM parameters are held fixed when training the agent.


\textbf{Training Stability.} Transformers have notably had stability issues when used in RL settings, especially in cases where the rewards are sparse.
GTrXL \citep{gtrxl}, in particular, adds GRU gating layers to try to alleviate this problem.
In TransDreamer, however, since the transformer parameters are held fixed during agent training and only trained to predict images, rewards, and discounts, we find that we do not encounter similar stability issues even without any additional gating.






\textbf{Prioritized Replay.} Since the training signal for agent training is based on only the rewards the agent receives, the reward prediction in the world model is especially important for learning a good agent.
Learning a good reward predictor, on the other hand, relies on the agent having a good enough policy so that it can collect trajectories with reward, especially in environments with sparse rewards.
To facilitate this, we optionally weight the replay buffer so that trajectories with higher rewards are more likely to be sampled.
In particular, we sample only from nonzero-reward trajectories $\alpha$-percentage of the time where $\alpha \in [0, 1]$.
The remaining trajectories are sampled uniformly across the replay buffer.


\textbf{Number of Imagination Trajectories.}
Due to the increased memory requirements of transformers compared with RNNs, we find that it is not feasible to generate imagined trajectories from every state sampled from the replay buffer, as is done in Dreamer.
Instead, we randomly choose a smaller subset of starting states of size $K$ to generate imagined trajectories from.
While this effectively reduces the number of trajectories the agent can learn from in any given iteration, we find that we are still able to achieve performance comparable or better than Dreamer.









\section{Related Works}
\textbf{Transformers in RL}.
Transformers have been used successfully in diverse domains including NLP \citep{transformer, bert, gpt1, gpt2, gpt3}, computer vision \citep{igpt, vit, imagetransformer}, and video generation \citep{vidtrans, vgpt}.
\citet{gtrxl} address the problem of using transformers in RL and showed that adding gating layers on top of the transformers layers can stabilize training.
Subsequent works addressed the increased computational load of using a transformer for an agent's policy \citep{rdn, ald}.
\cite{decision_transformer, tto} take a different approach by modeling the RL problem as a sequence modeling problem and use a transformer to predict actions without additional networks for an actor or critic.
Several recent works also explore long-term video generation with transformers which is related to building world models using transformer-based architectures~\citep{ocvt,oat}.




\textbf{Stochastic Transformers}.
Stochasticity has been added to several transformer-based architectures in the context of response generation \citep{variationaltrans}, sign language translation \citep{voskou2021stochastic}, story completion \citep{tcvae}, and layout generation \citep{variationaltrans2}.
\cite{smct} introduce the Sequential Monte Carlo Transformer which adds stochastic hidden states to the network architecture and outputs a distribution of predictions allowing for uncertainty quantification.
To our knowledge, no previous work investigates stochastic transformers in the context of world models and MBRL.

\textbf{Model-based RL}.
Dyna~\citep{dyna} introduced a general framework for MBRL that our model is based on.
SimPle~\citep{simple} adopts this framework by making predictions at the pixel level and training a PPO agent on that model.
Our work mainly builds off of the Dreamer~\citep{dreamer, dreamerv2} framework.
The RSSM is first introduced in PlaNet~\citep{planet} where it is used for planning in the latent space.
\cite{worldmodels} use a VAE with an RNN as the world model and learns a policy with an evolution strategy.
MuZero~\citep{muzero} uses task-specific rewards to build a model and Monte-Carlo Tree Search to solve RL tasks.

\section{Experiments}

In this section, we compare TransDreamer and Dreamer on a variety of tasks, from tasks 
that require long-term memory and reasoning to tasks that can be solved with only short-term memory.
We try to answer the following three questions:
1) How do TransDreamer and Dreamer perform in tasks that require long-term memory and reasoning?
2) How do the learned world models of TransDreamer and Dreamer compare?
3) Can TransDreamer also work comparably to Dreamer in environments that require short-term memory?

\begin{figure*}[t]
     \centering
     \begin{subfigure}[b]{0.32\textwidth}
         \centering
         \includegraphics[width=0.49\textwidth]{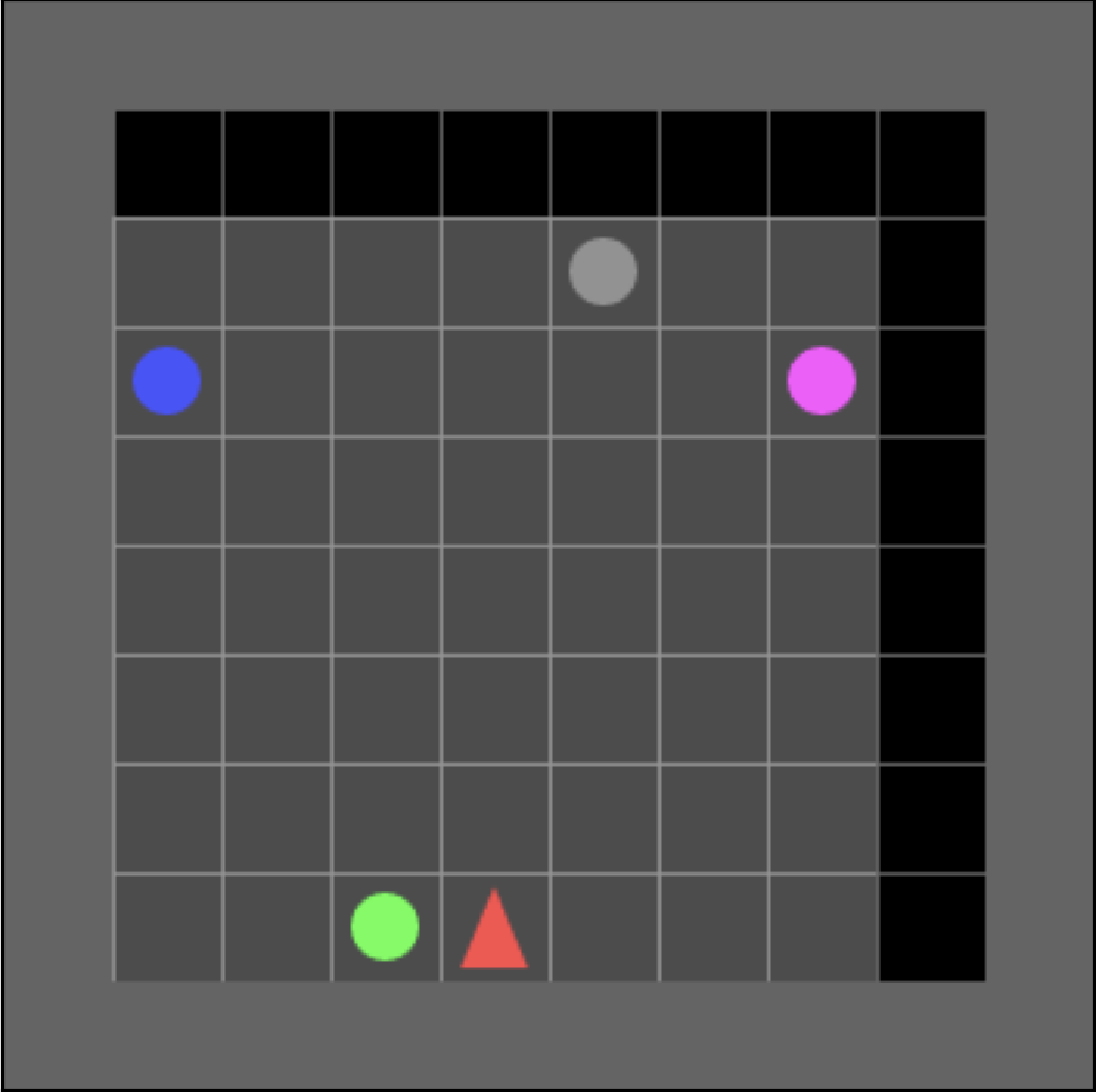}
         \includegraphics[width=0.49\textwidth]{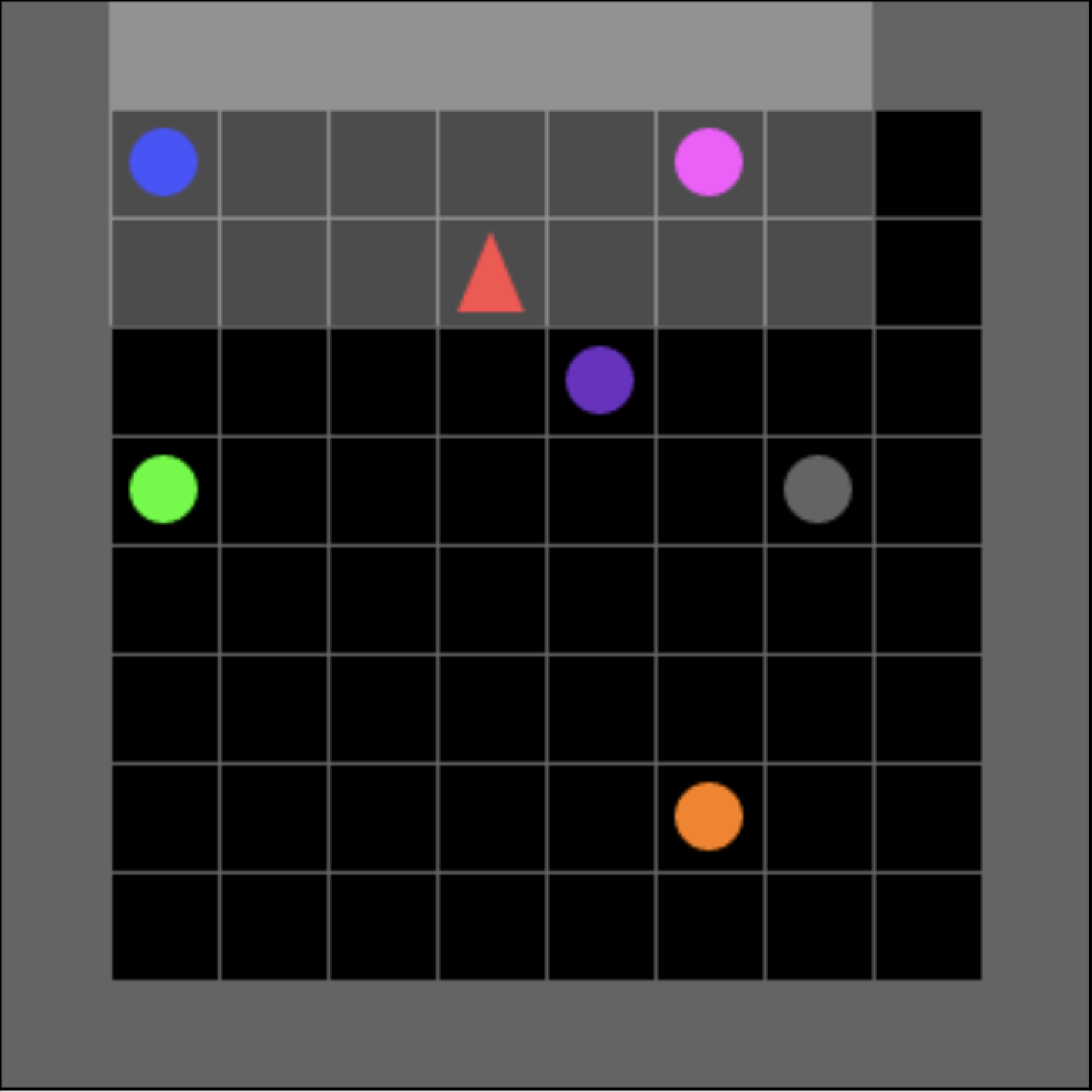}
         \caption{2D 4-Ball and 6-Ball}
         \label{fig:2d_env}
     \end{subfigure}
     \begin{subfigure}[b]{0.45\textwidth}
         \centering
         \includegraphics[width=0.32\textwidth]{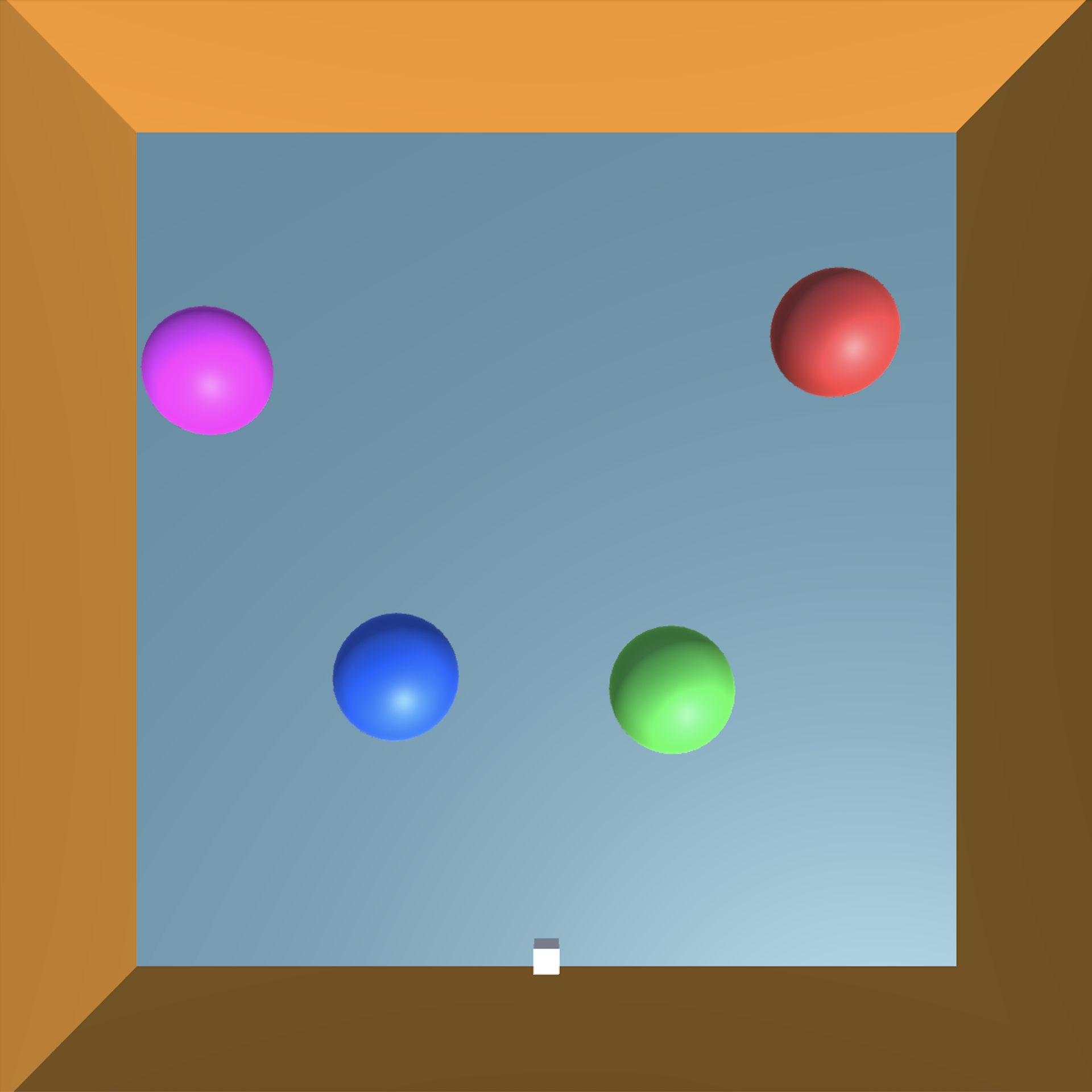}
         \includegraphics[width=0.32\textwidth]{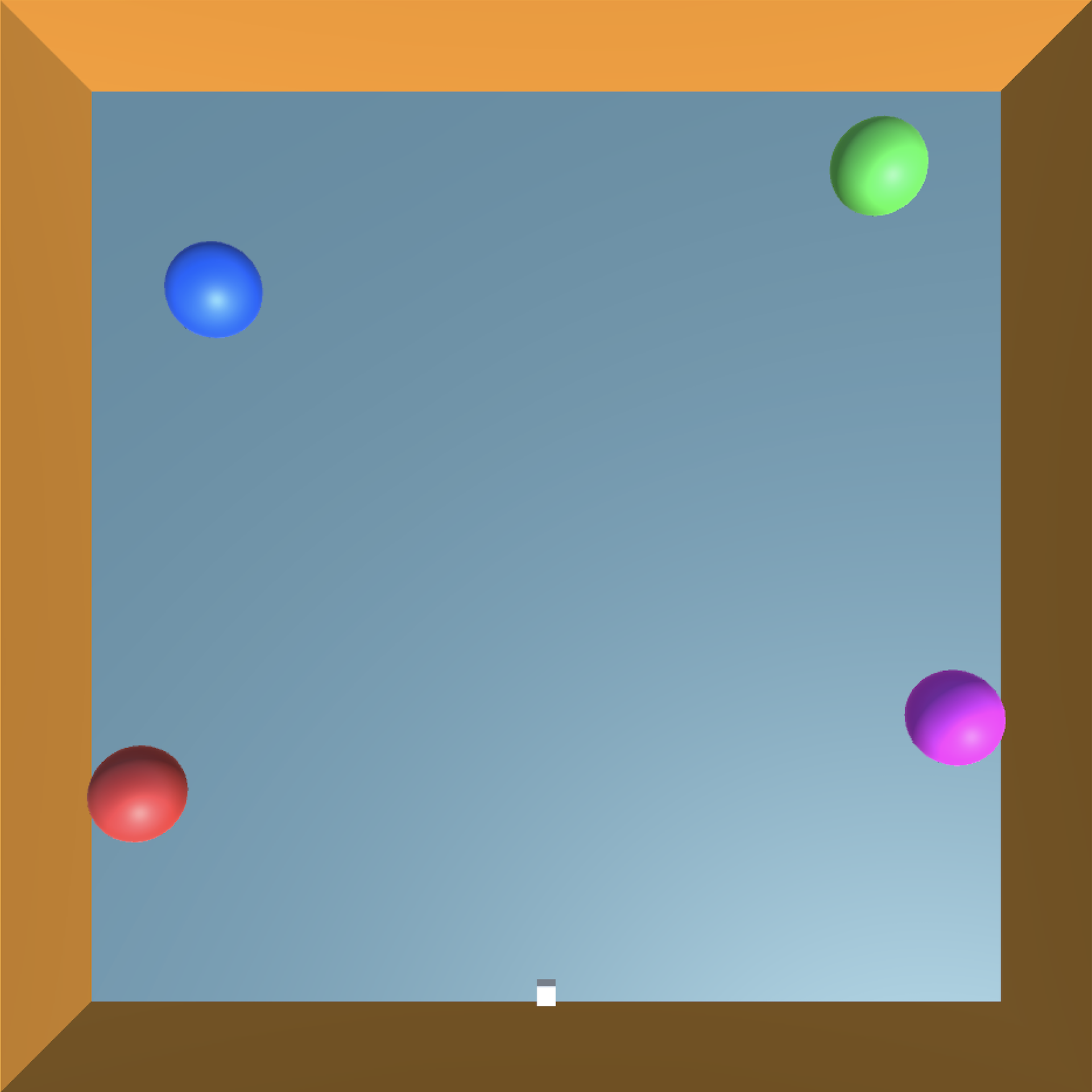}
         \includegraphics[width=0.32\textwidth]{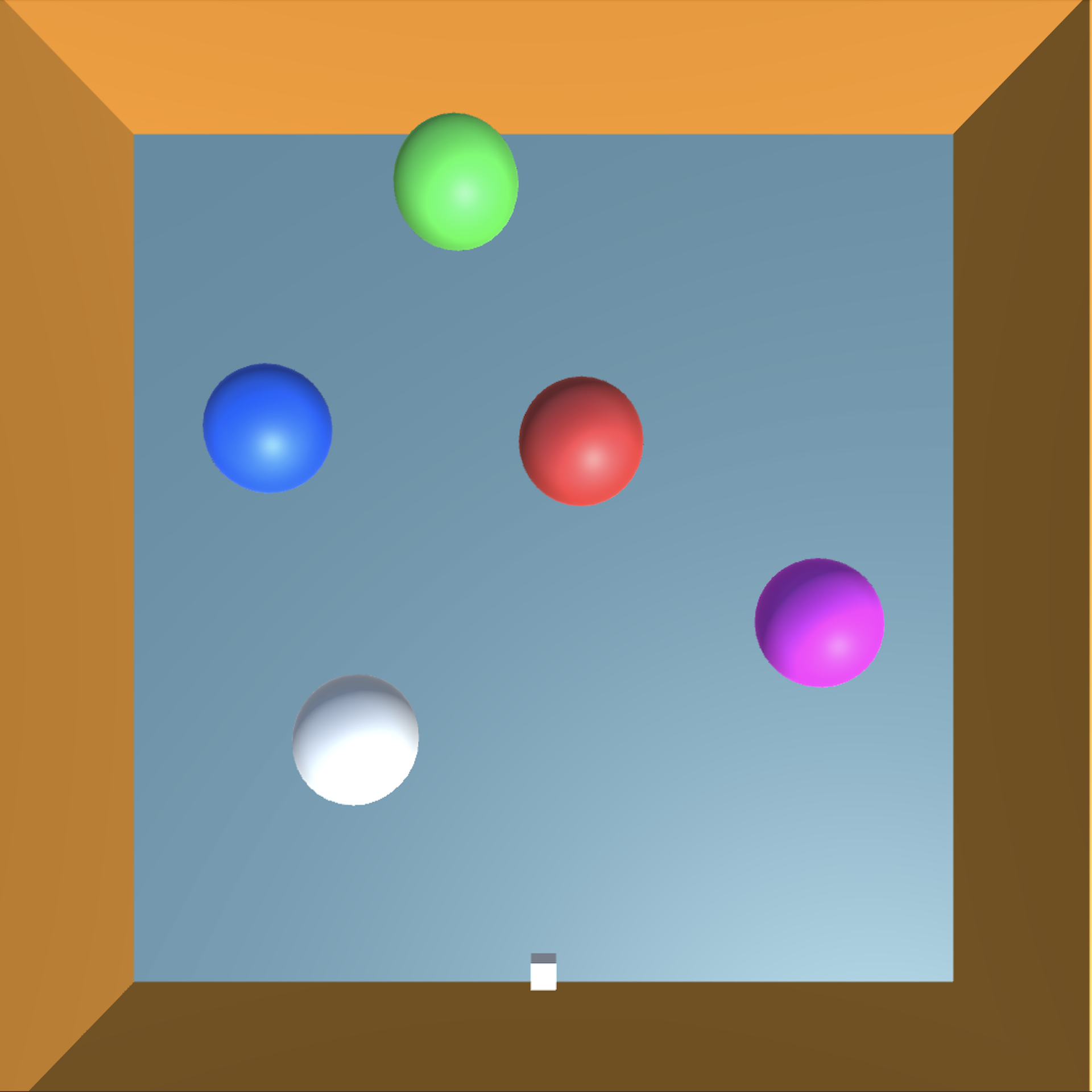}
         \caption{3D 4-Ball Dense/Sparse and 5-Ball Dense}
         \label{fig:3d_env_overview}
    \end{subfigure}
    \begin{subfigure}[b]{0.20\textwidth}
        \centering
         \includegraphics[width=0.72\textwidth]{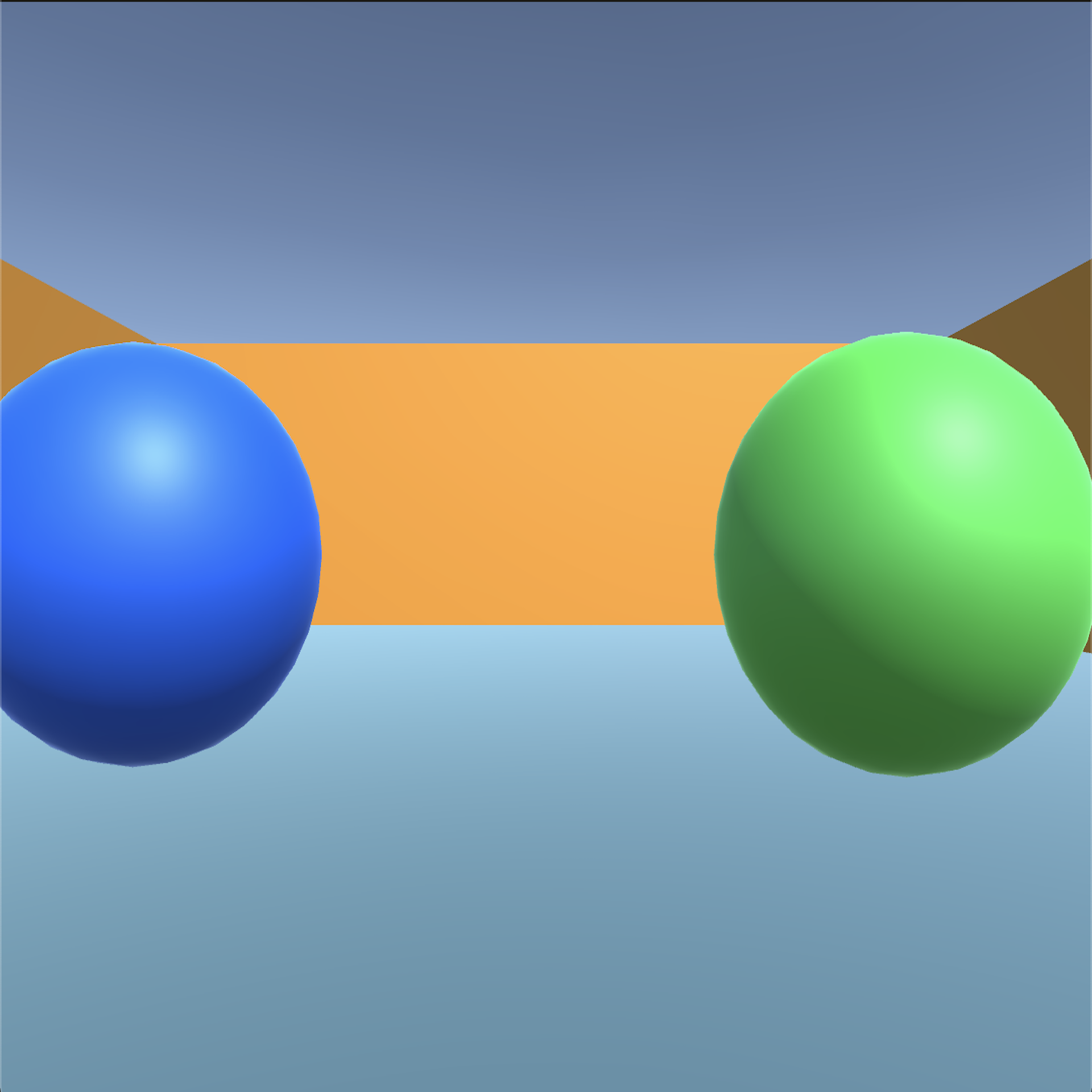}
         \caption{3D Agent View}
         \label{fig:3d_env_agent}
    \end{subfigure}
    \hfill
\vspace{-3mm}
\caption{Hidden Order Discovery Environments.}
\label{fig:env}
\vspace{-6mm}
\end{figure*}

To answer the first question, we created a new set of tasks called Hidden Order Discovery that is inspired by the Numpad task \citep{metarl_numpad_ref, gtrxl}.
We create both 2D and 3D versions of this task.
The 2D environment, built with the Minigrid \citep{gym_minigrid} framework, provides a top-down view of the agent navigating a room while the 3D environment, built with Unity \citep{unity}, provides a more partially observable and visually rich first-person view of the agent.
These tasks require long-term memory and reasoning to solve.
To answer the second question, we thoroughly analyze the quality of the world model that is learned in solving these tasks both quantitatively and qualitatively.
Lastly, to answer the third question, we compared TransDreamer and Dreamer on some tasks in DeepMind Control Suite (DMC) \citep{dmcontrol} and Atari \citep{atari_env}.
These tasks are almost fully observable and do not require long-term memory and reasoning to solve.

\subsection{Hidden Order Discovery in 2D Object Room}

To evaluate our model on the tasks that require long-term memory, we created a new task called Hidden Order Discovery inspired by the NumPad task \citep{metarl_numpad_ref, gtrxl}. 
In this task, there are several color balls in a 2D grid, as shown in Fig. \ref{fig:2d_env}.
The agent (illustrated by the red triangle) can only see the highlighted area in front of it, so this is a partially observable environment.
For each episode, there is a hidden order of balls and the agent must collect the balls \textit{in the correct order}.
If the agent fails, the map is reset, and the agent needs to start from the first ball.
Note that when the map is reset, the agent position and the hidden order are not changed but all the balls are reset to their initial positions. Therefore, to find the hidden ball order efficiently, the agent can benefit from remembering what it has tried in the past in the current episode.
When a ball is collected in the correct order, a reward of +3 is given, but if the map is reset due to the agent collecting the incorrect ball, the rewards for balls visited in previous tries are 0.
This prevents the agent from collecting a high reward from just constantly revisiting the first ball in the sequence.
When the agent successfully collects every ball in the hidden order, the map and rewards are reset.
The hidden order and ball positions are randomized per episode.

\begin{figure*}[t]
\centering
\includegraphics[width=0.95\linewidth]{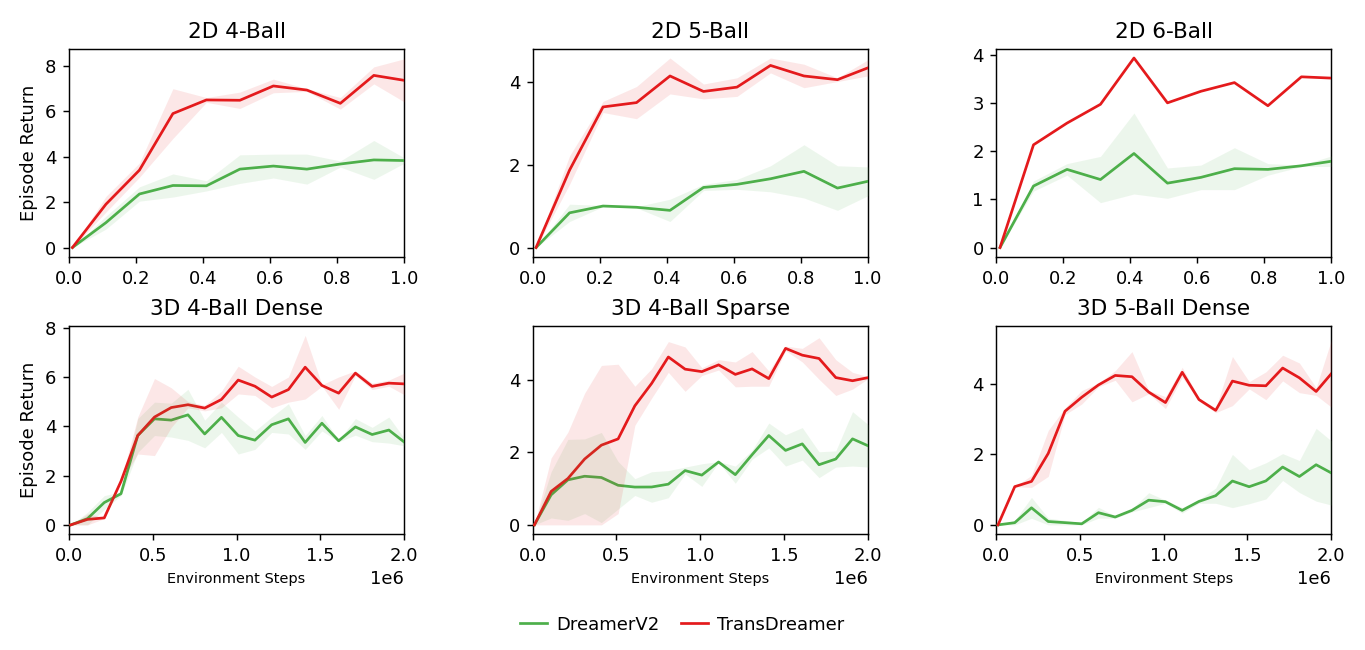}
\vspace{-3mm}
\caption{Comparison between DreamerV2 and TransDreamer in Hidden Order Discovery tasks}
\vspace{-3mm}
\label{fig:hod_result}
\end{figure*}

We evaluate in environments with 4, 5, and 6 balls where the grid size is 8x8 cells, and the agent is given 100 time steps to collect as much reward as possible.
The results are shown in Figure \ref{fig:hod_result}.
We see that TransDreamer outperforms Dreamer in all of these configurations. Since the reward for correctly collecting one ball is +3, an average reward of 3 means that on average the agent collects the first ball correctly, and an average reward of 6 means that on average the agent collects the first two balls correctly, and so on. For the 4-Ball configuration, TransDreamer reaches an episode reward of around 7 while Dreamer's episode reward is around 4.
This means that in the 4-Balls setting, TransDreamer averages collecting over two balls in the correct order, whereas Dreamer only collects one ball in the correct order. 

To obtain further understanding beyond the averaged behavior,
we also measure the success rate of each agent, defined as the percentage of trajectories where an agent collects \textit{all} balls in the correct order at least once.
For the 4-Ball configuration, we find that TransDreamer has a success ratio of 23\% and Dreamer has a success ratio of only 7\%, providing further evidence that TransDreamer can better solve this task than Dreamer.
A full comparison of the success ratio is reported in Appendix Table \ref{table:success_ratio}.
The difficulty of this task increases as the number of balls increases since with more balls, there are more combinations for the agent to try before determining the hidden order.
Thus, we see the performance for both degrade as the number of balls increases.

We emphasize the difficulty of this task.
Because the hidden order is randomized in each episode, in the worst-case scenario, discovering the first ball in the given order would require searching through all 4 balls.
Then, determining the second ball in the sequence would require searching among the remaining 3 balls, while always remembering what has happened before in order not to waste time by visiting balls already known to be incorrect.
The higher scores for TransDreamer provides some evidence that the transformer-based architecture is effective in tasks that require this long-term memory and reasoning.

\subsection{Hidden Order Discovery in 3D Object Room}

To evaluate this task in a more realistic environment, we also implemented a 3D version of Hidden Order Discovery in Unity \citep{unity}.
The reward structure is the same as the 2D task, but the agent view is a 3D first-person view.
Figure \ref{fig:3d_env_overview} shows an overview view of the different configurations and Figure \ref{fig:3d_env_agent} shows the agent's first-person view.
Compared to the 2D environment, since the environment is larger, it takes more steps to navigate to the balls, especially in the sparse setting.
Therefore, with the 3D environment, we can evaluate how TransDreamer can handle long-term dependency and complex reasoning more clearly.


We implemented 3 settings for this task.
The 4-Ball and 5-Ball Dense environments have 4 and 5 balls, separated by at least one ball-length each.
The 4-Ball Sparse environment tests longer-term memory by increasing the distance between balls to three ball-lengths so the agent needs to navigate a longer distance between balls.
The results are shown in Figure \ref{fig:hod_result}. Even if the 3D environment provides more severe partial observability and longer-term dependency than the 2D environment due to its degree of freedom in exploring a larger environment with a first-person view, we see that TransDreamer obtains similar outperforming results as we obtained in the 2D Object Room. 
Next, we compare the quality of the trajectories imagined by the TSSM and the RSSM by measuring the generation performance quantitatively and qualitatively.

\subsection{World Model - Quantitative Results}



We measure the Mean Squared Error (MSE) of the predicted images and the reward prediction accuracy during the action-conditioned generation in the 3D 5-Ball Dense configuration.
Even though the image is not directly used for policy training, the quality of the predicted image can serve as a proxy for measuring latent state prediction accuracy.
Reward prediction accuracy, on the other hand, is directly related to policy training, and may provide some insights into why TransDreamer performs better than Dreamer in the above environments.

\textbf{Image Generation.} For a fair comparison, we separately trained the TSSM and the RSSM with the same set of trajectories without any policy training.
Given a trajectory of 100 time steps, we measure the generation quality for several different context lengths and measure the per image MSE in the remaining generated steps.
This allows us to measure the generation quality given different amounts of historical context and analyze how the different models utilize this context.
The reported MSE is for the foreground objects (i.e., the balls), since that is where the most important information for this task is and more than 60\% of the gap in overall MSE can be attributed to the foreground, despite the balls only occupying a small portion of the image most of the time (see Appendix Table \ref{table:gen_mse_full} for overall MSE results).
The results are shown in Table \ref{gen_mse}.
We see that TransDreamer generally achieves lower or comparable MSE when compared with Dreamer.
As expected, more steps given in the context results in lower MSE since the agents have more opportunity to see the entire environment before making predictions.
In the 4-Ball Sparse setting, the MSE between TransDreamer and Dreamer are very comparable.
This may be because when the environment is sparse, the agent sees the foreground objects less frequently.

\textbf{Reward Prediction.} 
Since the reward is zero for most time steps and both models can predict zero reward time steps well (see Appendix Table \ref{table:reward_prediction_full}), we focus on the reward prediction accuracy for the nonzero +3 reward time steps.
Since the reward is a continuous value, in order to obtain an accuracy, we classify rewards as positive if they are predicted in the range $3 \pm 0.3$.
The results are shown in Table \ref{reward_prediction}.
We once again see that TransDreamer generally achieves more accurate reward prediction than Dreamer, with longer contexts resulting in higher accuracy.
For the 5-Ball Dense environments, however, Dreamer's reward prediction does not improve much as the context increases.
This can indicate that Dreamer is not fully taking advantage of the additional context when predicting rewards in these more complex settings.
TransDreamer, in contrast, does continue to improve when given more context, showing that the transformer architecture can take advantage of the increased context in making more accurate predictions. A full version including zero-reward prediction accuracy is reported in Appendix Table \ref{table:reward_prediction_full}.

\begin{table}[t]
    \caption{World Model Quantitative Comparison}
    \vspace{-2mm}
    \begin{subtable}{.46\linewidth}
		\caption{Image Generation Foreground MSE}
			\resizebox{\textwidth}{!}{\begin{tabular}{ccccc}
				\toprule
				 \multirow{2}{*}{Task}  & \multirow{2}{*}{Model} & \multicolumn{3}{c}{Context} \\  &&60 &70 &80\\ 
				\hline
				\multirow{2}{*}{4-Ball Dense} &TransDreamer  &\textbf{211.2} &\textbf{133.1} &\textbf{69.8}\\
				&DreamerV2 &281.9 &194.2  &110.8\\
				
				\hline
				\multirow{2}{*}{4-Ball Sparse} &TransDreamer &\textbf{195.5} &\textbf{115.2} &\textbf{56.8}\\
				&DreamerV2 &215.8 &138.6  &72.4\\
				
				\hline
				\multirow{2}{*}{5-Ball Dense} &TransDreamer &\textbf{245.2} &\textbf{163.1} &\textbf{85.0}\\
				&DreamerV2 &300.9 &217.0  &124.9\\
				\bottomrule
			\end{tabular}}
			\label{gen_mse}
    \end{subtable}%
    \hfill{}
    \begin{subtable}{.43\linewidth}
		\caption{Reward Prediction Accuracy}
		\resizebox{\textwidth}{!}{\begin{tabular}{ccccc}
				\toprule
				 \multirow{2}{*}{Task}  & \multirow{2}{*}{Model} & \multicolumn{3}{c}{Context} \\  &&60 &70 &80\\ 
				 \hline
				
				\multirow{2}{*}{4-Ball Dense} &Transdreamer  &\textbf{46.9}    &\textbf{53.2} &\textbf{73.2}  \\
				&DreamerV2 &28.2 &34.6 &50.5 \\
				
				\hline
				\multirow{2}{*}{4-Ball Sparse} &Transdreamer &\textbf{32.4} &\textbf{36.5} &\textbf{48.6} \\
				&DreamerV2 &32.0 &33.3 &42.3 \\
				
				\hline
				\multirow{2}{*}{5-Ball Dense} &Transdreamer  &\textbf{17.7} &\textbf{18.1} &\textbf{32.3} \\
				&DreamerV2 &9.8 &6.2 &15.3 \\
				\bottomrule
		\end{tabular}}
		\label{reward_prediction}
    \end{subtable} 
    \vspace{-2mm}
\end{table}

\begin{figure*}[t]
\centering
\begin{subfigure}[b]{\textwidth}
\centering
\includegraphics[width=\linewidth]{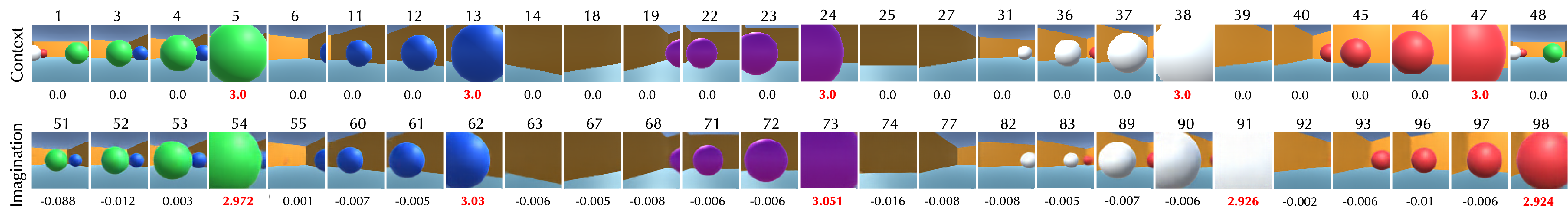}
\vspace{-6mm}
\caption{Imagined Trajectories from TransDreamer}
\end{subfigure}
\begin{subfigure}[b]{0.85\textwidth}
\centering
\includegraphics[width=\linewidth]{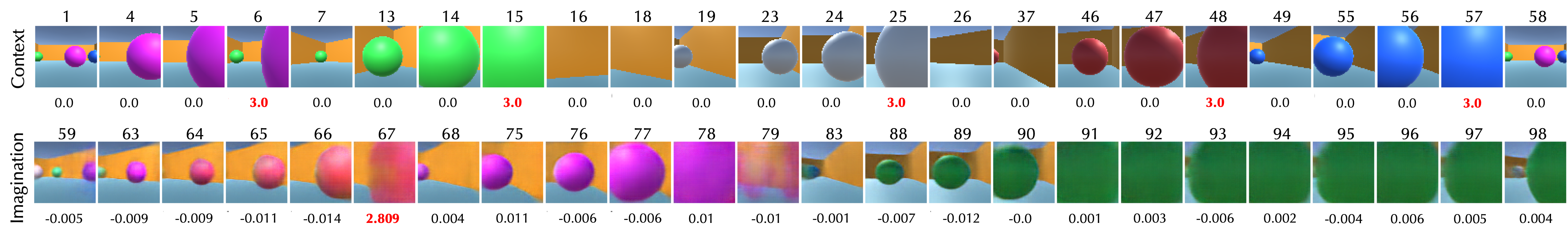}
\vspace{-6mm}
\caption{Imagined Trajectories from Dreamer}
\end{subfigure}
\vspace{-3mm}
\caption{Imagined trajectories comparison between DreamerV2 and TransDreamer
}
\label{fig:imagination}
\vspace{-6mm}
\end{figure*}

\subsection{World Model - Qualitative Comparison}

In Figure \ref{fig:imagination}, we show the imagined trajectories from TransDreamer and Dreamer in the 5-Ball Dense environment.
We provide context timesteps from each agent's trained policy up to when all the balls are collected for the first time.
After this, the agent and balls reset to their original positions (frame 48 for TransDreamer and frame 58 for Dreamer).
We then imagine the rest of the trajectory up to 100 total steps for each agent.
Since the context timesteps contain information about the correct order of balls, ideally the agent would revisit the balls in this order during the imagination timesteps and correctly predict the rewards when the balls are collected.
Note that the context frames for Dreamer and TransDreamer are different since they are based on trajectories from their own policies.
This is necessary because the world model is trained from the trajectories of each agent's policy.
Providing context that is not from the agent's policy would not necessarily be in the training distribution of the respective world model.
See Figure \ref{fig:imagination_ood} in the Appendix for an example of when the same context is given to both agents.
Despite this, however, we can still see some clear differences between the quality of the imagination steps as well as the reward predictions.

In particular, TransDreamer is able to predict the appearance of the environment correctly as well as the collection and subsequent disappearance of the balls in frames 54, 62, 73, 90, and 98.
It also accurately predicts the reward at these timesteps of around +3 (highlighted in red).
For timesteps where there is no reward, it correctly predicts a reward around 0.
Dreamer, on the other hand, predicts images that are blurrier than TransDreamer.
Furthermore, the imagined trajectories are incorrect.
While it does predict the collection of the red ball and the reward in frame 67, this color is incorrect since the first ball should be purple.
When it subsequently predicts the collection of the purple ball in frame 78, it is again the wrong color and no reward is predicted.
This error seems to compound as the dark green ball it predicts at the end of the trajectory is not even one of the possible colors in the environment.
This shows that the quality of the world model is better in TransDreamer than Dreamer, especially in the later steps of imagination where the long-term memory is more important.
This can be a reason why TransDreamer outperforms Dreamer in these tasks.

\subsection{Short-Term Memory Tasks in DMC and Atari} \label{sec:dmc_atari}

As the final validation, we perform a sanity check by testing the proposed model on a few simple DMC and Atari tasks. We note that it is expected that these tasks may favor Dreamer over TransDreamer, because solving these tasks does not require long-term and complex memory interactions, but modeling the dynamics of just the last few steps can suffice\footnote{Some Atari games require long-term memory along with a good exploration policy. We do not choose these games as we do not address the exploration problem in this work.}.
Specifically, we expect that RNN-based models can learn \textit{faster} than transformer-based models because the former has the specific inductive bias to focus on the near-term history. 
Nevertheless, TransDreamer is supposed to converge eventually to an accuracy  similar to that of Dreamer, and it is an important step to see whether these expectations are met.

We follow the configurations used by the authors in Dreamer \citep{dreamer} for DMC and DreamerV2 \citep{dreamerv2} for Atari.
The only difference is that Dreamer uses the imagined trajectory from all states sampled from the replay buffer, whereas TransDreamer uses a few randomly selected states. Configurations for Transformer are described in Appendix \ref{sec:appx_dmc_atari}.
As shown in Fig. \ref{fig:dmc_result}, Dreamer and TransDreamer eventually reach comparable performance as expected, but TransDreamer is slower to saturate in general than Dreamer except for a few tasks.
Interestingly, TransDreamer shows slightly better performance and faster convergence for the DMC Cheetah Run.


\begin{figure*}[t]
\centering
\includegraphics[width=0.95\linewidth]{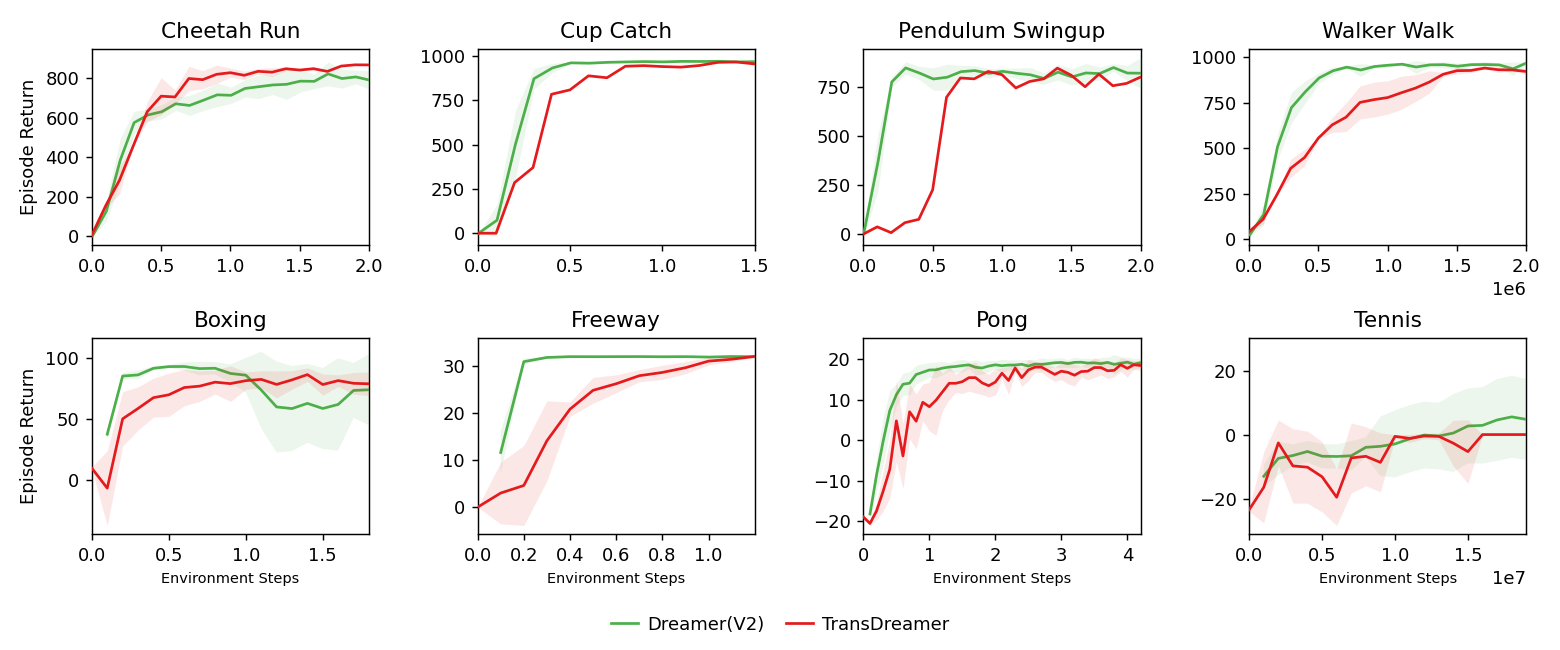}
\vspace{-3mm}
\caption{Comparison between Dreamer and TransDreamer on a few DMC (upper row) and Atari tasks (bottom row) for short-term memory test. As expected, TransDreamer converges to a similar return value but slowly.
}
\label{fig:dmc_result}
\vspace{-6mm}
\end{figure*}

\section{Conclusion}

We proposed TransDreamer, a transformer-based MBRL agent, and the Transformer State-Space Model (TSSM), the first transformer-based stochastic world model.
TransDreamer shows comparable performance with Dreamer on DMC and Atari tasks that do not require long-term memory, and outperforms Dreamer on Hidden Order Discovery tasks that require long-term complex memory interactions.
We also show that image generation and reward prediction of TSSM is better than Dreamer qualitatively and quantitatively.
Future work may involve validating our model on more complex tasks such as Crafter \citep{crafter}.

\section*{Acknowledgments}

This work is supported by Brain Pool Plus Program (No. 2021H1D3A2A03103645) through the National Research Foundation of Korea (NRF) funded by the Ministry of Science and ICT.

\bibliography{refs, refs_ahn}
\bibliographystyle{iclr2022_conference}

\newpage
\appendix
\section{Appendix}

\subsection{DreamerV2} \cite{dreamerv2} makes several additional changes to the framework that are found to improve performance on the Atari environment. First, instead of using a continuous stochastic hidden state, a discrete state is used. Second, straight-through gradients~\citep{straight-through} are used to differentiate through the discrete states and actions. Due to the bias induced by the straight-through estimator, REINFORCE gradient or a mixed gradient of REINFORCE and the dynamics backpropagation is used. Lastly, they use KL balancing, separately scaling the prior cross entropy and the posterior entropy in the KL loss.

\subsection{TransDreamer Loss Function} \label{appx:sec_loss_func}

We optimize the following objective, which is the negative ELBO of the action conditioned model with additional terms for predicting the reward and discount,
\begin{align}
\begin{split}
\mathcal{L}_\text{TSSM}(\phi) = \sm{t}{T} \Bigg( &\eE_{\pd{\tau}{t}q_{\phi(z_\tau|x_\tau)}}[-\eta_x \ln p_\phi(x_t|h_t, z_t) -\eta_r \ln p_\phi(r_t|h_t, z_t) -\eta_\gamma \ln p_\phi(\gamma_t|h_t, z_t)] \\
&+ \eE_{\pd{\tau}{t-1}q_\phi(z_\tau|x_\tau)} \left[ \KL\left[ q_\phi(z_t|x_t) \parallel p_\phi(z_t|z_\ottmo, a_\ottmo) \right]\right]\Bigg).\nn
\end{split}
\end{align}
Here, $\eta_x$, $\eta_r$, and $\eta_\gamma$ are hyperparameters used to scale the loss terms. The derivation of the ELBO can be found in the below.

\subsubsection{ELBO} The generative model is $p(o_t, z_\otT| a_\otT) = \prod_t p(o_t|h_t,z_t)p(z_t|z_\ottmo, a_\ottmo)$ where $o_t=(x_t,r_t,\gamma_t)$ and $h_t = f_\text{transformer}(z_{1:\tmo}, a_{1:\tmo})$. By approximating the posterior by $q(z_t|x_t)$, a variational posterior is $q(z_\otT | o_\otT, a_\otT) = \prod_t q(z_t|x_t)$. By the importance weighting and Jensen's inequality, we can write as follows:
\eq{
\ln p(o_\otT | a_\otT) 
&= \ln \eE_{p(z_\otT|o_\otT,a_\otT)} \left[ \pd{t}{T}p(o_t|h_t,z_t)\right]\nn \\
&= \ln \eE_{q(z_\otT|o_\otT,a_\otT)} \left[ \pd{t}{T}p(o_t|h_t,z_t) p(z_t|z_\ottmo,a_\ottmo) / q(z_t|x_t)\right]\nn \\
&\ge \eE_{\pd{t}{T}q(z_t|x_t)} \left[ \sm{t}{T} \ln p(o_t|h_t,z_t) + \ln p(z_t|z_\ottmo,a_\ottmo) - \ln q(z_t|x_t)\right]\nn\\
&= \sm{t}{T} \Bigg( \eE_{\pd{\tau}{t-1}q(z_\tau|x_\tau)}[\ln p(o_t|h_t,z_t)] \nn\\
&\hspace{5mm} - \eE_{\pd{\tau}{t-1}q(z_\tau|x_\tau)} \left[ \KL\left[ q(z_t|x_t) \parallel p(z_t|z_\ottmo, a_\ottmo) \right]\right]\Bigg)\\
&= \sm{t}{T} \Bigg( \eE_{\pd{\tau}{t-1}q(z_\tau|x_\tau)}[\ln p(x_t|h_t,z_t)+\ln p(r_t|h_t,z_t) + \ln p(\gamma_t|h_t,z_t)] \nn\\
&\hspace{5mm} - \eE_{\pd{\tau}{t-1}q(z_\tau|x_\tau)} \left[ \KL\left[ q(z_t|x_t) \parallel p(z_t|z_\ottmo, a_\ottmo) \right]\right]\Bigg)
}
where $p(o_t|h_t,z_t)=p(x_t|h_t,z_t)p(r_t|h_t,z_t)p(\gamma_t|h_t,z_t)$.

\begin{figure}[t]
     \centering
      \begin{subfigure}[b]{\textwidth}
          \centering
          \includegraphics[width=0.7\linewidth]{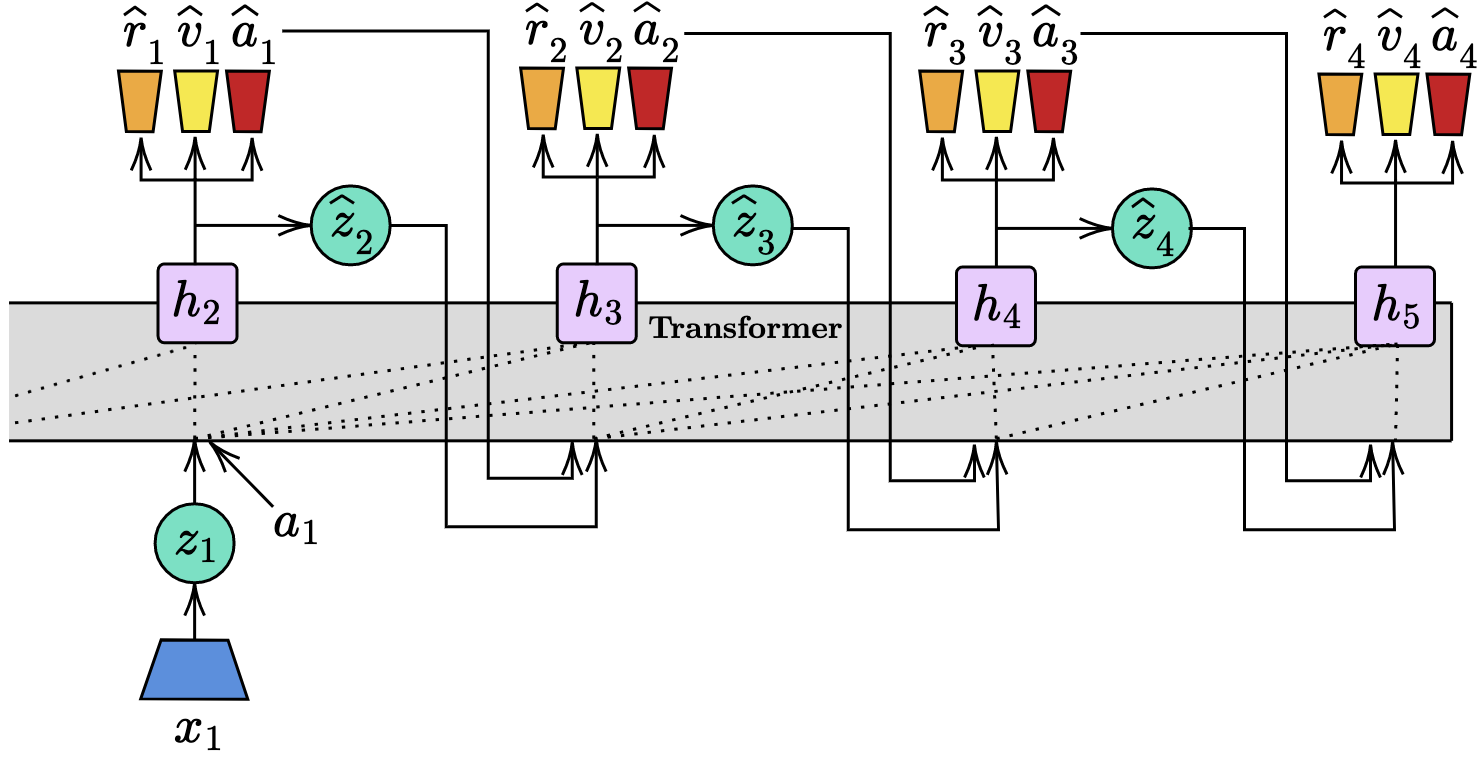}
      \end{subfigure}
      \caption{Transfomer-Based Trajectory Rollout for Actor Critic Learning.}
      \label{fig:rollout}
     \vspace{-3mm}
 \end{figure}
 
\subsection{DMC and Atari} \label{sec:appx_dmc_atari}
 
As written in Sec. \ref{sec:dmc_atari}, we used almost identical configurations with Dreamer and DreamerV2 by referring to the configuration file in \href{https://github.com/danijar/dreamerv2}{https://github.com/danijar/dreamerv2} (e.g., action repeat and training World Model and policy every 5 steps for DMC).
The one configuration we did modify is the number of imagined trajectories, which is not configurable in Dreamer, but is necessary in TransDreamer because imagining from every state in the batch requires too much computational resources.
We control this through a hyperparamter that limits the number of imagined trajectories per training sample.
For DMC and Atari, we use 3 imagined trajectories per sample.

Several other hyperparmeters are specific to the TSSM.
These include: whether or not to use gating or identity map reording as is done in GTrXL \citep{gtrxl}, the number of layers and heads to use for the transformer, the size of the hidden state for the MLP in the transformer, and whether or not to use relational positional embedding \citep{trxl}.
For DMC and Atari, we generally use 2-layer Transformer \citep{transformer} without dropout, gating, or identity map reordering.
One exception is for Atari Pong where we did find identity map reordering performed better.
We use 10 heads in the Multihead Attention and the dimensions of the hidden state for the MLP and Attention are 200 for DMC and 600 for Atari.
These are the same as the dimensions used in the deterministic state in DreamerV2.


\subsection{Hidden Order Discovery} \label{sec:appx_hod}

For 2D and 3D Hidden Order Discovery tasks, we measure the model in two aspects, the ability to deal with complex memory-based reasoning and the ability to extract long-term knowledge. Thus we design tasks either with an increased number of objects or the distances between any two objects or both. Specifically, on 2D tasks, we increase the number of balls from 4 to 6, while not changing the distance. The distance here is measured as the number of cells between any two balls. We sum the absolute difference along the $x$-axis and $y$-axis as the distance between any two balls. To control the long-term dependency, a threshold of 2 is applied to the distance of balls, i.e. the minimum distance between any two balls should not be less than 2 cells. For 3D tasks, we tested not only the reasoning complexity but also the ability to capture long-term dependency. For reasoning complexity, we compare 5-Ball Dense with 4-Ball Dense. For long term dependency, we compare 4-Ball sparse against 4-Ball Dense. The sparse setting has a larger distance between any two balls. We use the Euclidean distance as a measure of distance. Any two balls have a distance at least 4 units in the sparse setting, while in the dense setting, it is 2 units. 1 unit equals 1 ball size (diameter). Thus, in sparse setting, the distance between any two balls is at least 3 ball-size. For each task, the maximum steps for an episode is set as 100.

We implemented a 6-layer transformer with identity map reordering as TSSM for both 2D and 3D Hidden Order Discovery tasks. During imagination, only one state was randomly sampled as the starting state for imagination. We imagined till the trajectory's max step was reached. Empirically we found concatenating the intermediate output of attention layers together as $h_t$ accelerates the converge speed, so we applied this during experiments. Other hyperparameter configurations are kept the same as DreamerV2 crafter configuration, see https://github.com/danijar/crafter/issues/1 for details. For DreamerV2, we use the same configuration as DreamerV2 for crafter, except that we imagined 30 steps for agent learning.


As explained in the paper, to help train the world model better, we used a prioritized replay buffer for Dreamer and TransDreamer with $\alpha=0.5$. The sampling probability for each trajectory is set at the return of this trajectory divided by the overall return of the whole data buffer collected till that time, thus a trajectory with higher rewards will have a higher chance to be sampled. The rest $50\%$ of the batch are sampled uniformly from the whole data buffer.

Table \ref{table:success_ratio} shows the ratio of successfully completing at least one round of the hidden order on all the 2D and 3D tasks. We deploy the well-trained agent for each task to collect 1000 trajectories and count the ratio in the whole trajectories. As we can see, TransDreamer performs better than Dreamer by this metric. Note that the episode length is limited to 100, and succeeding in this task in 100 steps is difficult. For example, in the 4-Ball case, the chance of guessing the right order is 0.04 (1 over $4!$), and the agent needs to start collecting from the first ball when it collects an incorrect ball. Therefore the agent needs to start over again and again during exploration. When increasing the number of balls from 4 to 5, the chance of randomly guessing the order decreases by a factor of 5. We can observe this relation approximately on TransDreamer's performance, $23\% \to 5\%$, while Dreamer nearly fails.
\begin{table}[h]
	\begin{center}
		\caption{Success Rate for Complete Order Visitation}
		\vspace{-2mm}
		\begin{tabular}{ccccccc}
				\toprule
				\multirow{2}{*}{Task}   &\multicolumn{3}{c}{2D Object Room} &\multicolumn{3}{c}{3D Object Room} \\
								\cmidrule(lr){2-4} \cmidrule(lr){5-7}
				&4-Ball & 5-Ball & 6-Ball &4-Ball Dense & 4-Ball Sparse & 5-Ball Dense\\ 
				\hline
				TransDreamer  &\textbf{23\%} &\textbf{5\%} &\textbf{1\%} &\textbf{18\%} &\textbf{11\%} &\textbf{4\%}\\
				DreamerV2  & 7\% & 0\% & 0\% & 10\% & 1\% &0 \%\\
				
				\bottomrule
		\end{tabular}
		\label{table:success_ratio}
	\end{center}
	\vspace{-4mm}
\end{table}

\subsubsection{Full Quantitative Results}

\begin{wrapfigure}{r}{0.5\textwidth}
    \centering
    \includegraphics[width=0.5\textwidth]{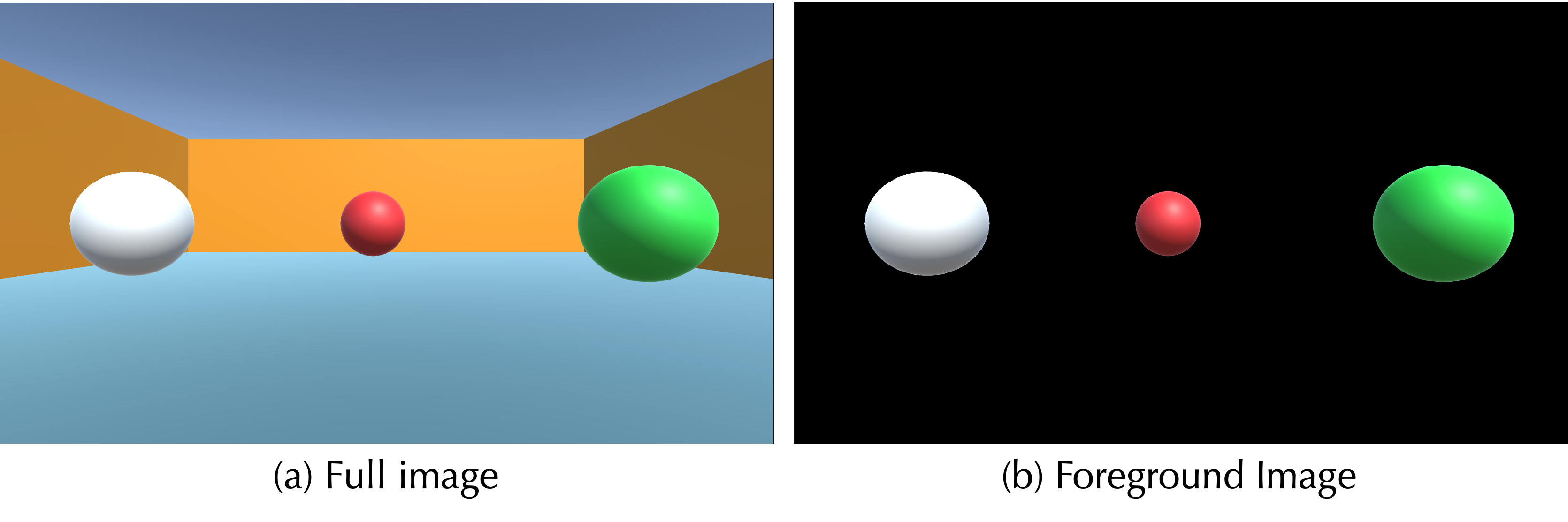}
    \caption{Image from Unity Foreground Camera}
    \vspace{-3mm}
    \label{fig:unity_fg}
\end{wrapfigure}
To compute the foreground MSE, we use Unity \citep{unity} to render a foreground image, Figure \ref{fig:unity_fg}, and from the foreground image, we infer a binary foreground mask to filter out the background from the predicted image.
The full MSE result is reported in Table \ref{table:gen_mse_full}. As can be seen, more than half of the overall MSE gap between TransDreamer and Dreamer happens in the foreground. For example, in the 4-Ball Dense, 60 context setting, the overall MSE gap between TransDreamer and Dreamer is 119.8, while 70.7 of the error occurs in the foreground.

Table \ref{table:reward_prediction_full} shows the reward prediction accuracy on both zero-reward and nonzero-reward timesteps for the 3D tasks. As mentioned in the paper, to measure prediction accuracy for +3 reward case, we classify it by labeling $3 \pm 0.3$ as positive. For 0 reward case, we classify it by labeling $\pm 0.01$ as positive.  We can see that TransDreamer outperforms Dreamer by a large gap on nonzero-reward in the 4-Ball Dense and the 5-Ball Dense. Both models perform well generally on zero-reward. In 4-Ball Sparse setting, The gap is smaller, we hypothesis that it is because in the sparse setting, the foreground balls are seen less frequently. 

\begin{table}[h]
	\begin{center}
		\caption{Image Generation MSE}
			\resizebox{\textwidth}{!}{\begin{tabular}{cccccccc}
				\toprule
				 \multirow{2}{*}{Task}  & \multirow{2}{*}{Model}  &\multicolumn{2}{c}{60 contexts / 40 targets} &\multicolumn{2}{c}{70 contexts / 30 targets} &\multicolumn{2}{c}{80 contexts / 20 targets}\\ 
				\cmidrule(lr){3-4} \cmidrule(lr){5-6} \cmidrule(lr){7-8}
				& &Overall & Foreground  &Overall & Foreground  &Overall & Foreground\\ \hline

				\multirow{2}{*}{4-Ball Dense} &TransDreamer &\textbf{458.0} &\textbf{211.2} &\textbf{281.9} &\textbf{133.1}  &\textbf{146.0} &\textbf{69.8}\\
				&DreamerV2 																	&577.8 &281.9			   &380.0              &194.2               &206.2              &110.8\\

				\hline
				\multirow{2}{*}{4-Ball Sparse} &TransDreamer &\textbf{448.8} &\textbf{195.5} &\textbf{261.4} &\textbf{115.2}  &\textbf{128.1} &\textbf{56.8}\\
				&DreamerV2 									&462.6          &215.8          &279.7          &138.6               &145.1              &72.4\\

				\hline
				\multirow{2}{*}{5-Ball Dense} &TransDreamer &\textbf{516.0} &\textbf{245.2} &\textbf{329.9} &\textbf{163.1}  &\textbf{167.4} &\textbf{85.0}\\
				&DreamerV2 													  &605.1              &300.9               &413.8              &217.0              &231.6              &124.9\\
				\bottomrule
			\end{tabular}}
			\label{table:gen_mse_full}
	\end{center}
	\vspace{-4mm}
\end{table}

\begin{table}[h]
	\begin{center}
		\caption{Reward Prediction Accuracy}
		\resizebox{\textwidth}{!}{\begin{tabular}{cccccccc}
				\toprule
				\multirow{2}{*}{Task}  & \multirow{2}{*}{Model}  &\multicolumn{2}{c}{60 contexts / 40 targets} &\multicolumn{2}{c}{70 contexts / 30 targets} &\multicolumn{2}{c}{80 contexts / 20 targets}\\ 
				\cmidrule(lr){3-4} \cmidrule(lr){5-6} \cmidrule(lr){7-8}
				& &Zero & Non-zero  &Zero & Non-zero  &Zero & Non-zero\\ 			\hline

				\multirow{2}{*}{4-Ball Dense} &Transdreamer &\textbf{94.9} &\textbf{46.9}  &\textbf{94.7}  &\textbf{53.2}  &\textbf{95.4} &\textbf{73.2}  \\
				&DreamerV2 &93.7 &28.2 &93.6 &34.6 &94.2 &50.5 \\

				\hline
				\multirow{2}{*}{4-Ball Sparse} &Transdreamer &\textbf{96.4} &\textbf{32.4} &96.0 &\textbf{36.5} &\textbf{96.6} &\textbf{48.6} \\
				&DreamerV2                                  &95.6           &32.0 &\textbf{96.2} &33.3 &95.5 &42.3 \\

				\hline
				\multirow{2}{*}{5-Ball Dense} &Transdreamer &\textbf{92.5}  &\textbf{17.7} &\textbf{93.2} &\textbf{18.1} &\textbf{93.3} &\textbf{32.35} \\
				&DreamerV2 	                                                 &91.1 &9.8 &92.3 &6.2 &92.4 &15.3 \\
				\bottomrule
		\end{tabular}}
		\label{table:reward_prediction_full}
	\end{center}
	\vspace{-4mm}
\end{table}

\newpage
\subsection{World Model Imagination with Same Context}
\begin{figure*}[t]
\centering
\includegraphics[width=\linewidth]{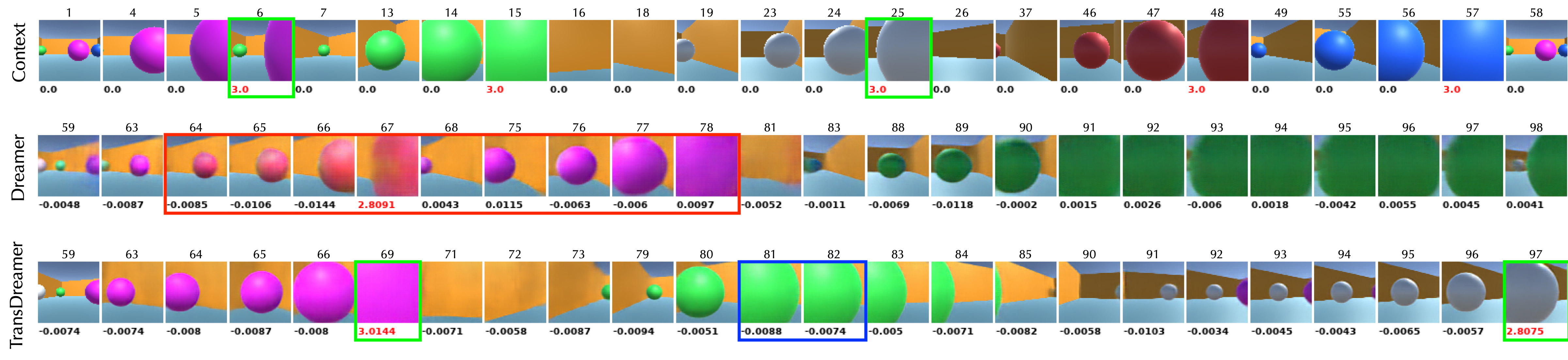}
\vspace{-7mm}
\caption{Imagined trajectories comparison between DreamerV2 and TransDreamer given same context
}
\label{fig:imagination_ood}
\end{figure*}

Different from Figure \ref{fig:imagination}, in Figure \ref{fig:imagination_ood}, we illustrated imagined trajectories from TransDreamer and Dreamer given the same contexts. This 5-Ball Dense sample is collected from Dreamer's agent learning process, so for TransDreamer, it is an out of distribution sample.
This is the same context given to Dreamer in Figure \ref{fig:imagination}.
Despite being an out of distribution sample, TransDreamer can still correctly imagine the balls and predicts rewards for the purple and white balls (Green box) correctly.
However, it does make a mistake predicting the reward for the green ball (blue box).
Nevertheless, even in this setting, the quality of imagination in TransDreamer is better than Dreamer.

\end{document}